\documentclass[conference, a4paper]{IEEEtran}
\IEEEoverridecommandlockouts

\usepackage{cite}
\usepackage{amsmath,amssymb,amsfonts}
\usepackage{algorithmic}
\usepackage{graphicx}
\usepackage{textcomp}
\usepackage{xcolor}
\def\BibTeX{{\rm B\kern-.05em{\sc i\kern-.025em b}\kern-.08em
    T\kern-.1667em\lower.7ex\hbox{E}\kern-.125emX}}
\usepackage[a4paper, left=1.45cm, right=1.5cm, top=2.0cm, bottom=4.5cm]{geometry}
\usepackage[colorlinks=true, urlcolor=blue, linkcolor=red]{hyperref}

\begin{document}

\title{3D Multi-Object Tracking Employing MS-GLMB Filter for Autonomous Driving
\thanks{This work was supported by Culture, Sports and Tourism R\&D Program through the Korea Creative Content Agency grant funded by the Korea government (MCST) in 2024 (R2022060001), and GIST-MIT Research Collaboration Project.}
}

\author{\IEEEauthorblockN{Linh Van Ma\IEEEauthorrefmark{1}, Muhammad Ishfaq Hussain\IEEEauthorrefmark{1}\IEEEauthorrefmark{3}, Kin-Choong Yow\IEEEauthorrefmark{2}, Moongu Jeon\IEEEauthorrefmark{1}}
\IEEEauthorblockA{\IEEEauthorrefmark{1}\textit{School of Electrical Engineering and Computer Science, GIST, Republic of Korea}\\
\IEEEauthorrefmark{2}\textit{Faculty of Engineering and Applied Science, University of Regina, Canada}\\
\IEEEauthorrefmark{3}\textit{Large-scale AI Research Group}, \textit{Korea Institute of Science and Technology Information (KISTI), South Korea} \\
\IEEEauthorrefmark{1}\{linh.mavan, ishfaqhussain, mgjeon\}@gist.ac.kr, \IEEEauthorrefmark{2}Kin-Choong.Yow@uregina.ca, \IEEEauthorrefmark{3}ishfaq@kisti.re.kr}}


\maketitle

\begin{abstract}
The MS-GLMB filter offers a robust framework for tracking multiple objects through the use of multi-sensor data. Building on this, the MV-GLMB and MV-GLMB-AB filters enhance the MS-GLMB capabilities by employing cameras for 3D multi-sensor multi-object tracking, effectively addressing occlusions. However, both filters depend on overlapping fields of view from the cameras to combine complementary information. In this paper, we introduce an improved approach that integrates an additional sensor, such as LiDAR, into the MS-GLMB framework for 3D multi-object tracking. Specifically, we present a new LiDAR measurement model, along with a multi-camera and LiDAR multi-object measurement model. Our experimental results demonstrate a significant improvement in tracking performance compared to existing MS-GLMB-based methods. Importantly, our method eliminates the need for overlapping fields of view, broadening the applicability of the MS-GLMB filter. Our source code for nuScenes dataset is available at \href{https://github.com/linh-gist/ms-glmb-nuScenes}{https://github.com/linh-gist/ms-glmb-nuScenes}.
\end{abstract}


\section{Introduction}

3D multi-object tracking is critical for autonomous driving, which has the potential to transform urban landscapes and save lives. Its goal is to determine the position, orientation, and size of objects in the environment over time. Detecting and tracking surrounding agents is vital for safe navigation. By considering temporal data, a tracking module can eliminate outliers from detection and effectively handle partial or complete occlusions. To accomplish this, modern self-driving cars utilize multiple sensors, along with advanced detection and tracking algorithms \cite{van2024visual}. This allows them to map the paths of various moving objects, such as pedestrians, cyclists, and vehicles. These trajectories can then be analyzed to understand motion patterns and driving behaviors, enhancing predictive capabilities and aiding in autonomous driving planning \cite{hussain2023artificial}.

Cameras provide precise measurements of edges, colors, and lighting, aiding in image plane classification and localization, though determining 3D positions from images is difficult. LiDAR point clouds offer less semantic detail but excel in accurate 3D localization, with reflectance being a key feature, despite being sparse and limited to a range of 50-150m. Although radar sensors extend the range to 200-300m, their observations are even sparser and less precise in localization compared to LiDAR \cite{hussain2022drivable}. Therefore, it is essential to develop methods to fuse information from multiple sources to enhance autonomous driving capabilities.

\begin{figure*}[ht]
\begin{centering}
\includegraphics[width=\textwidth]{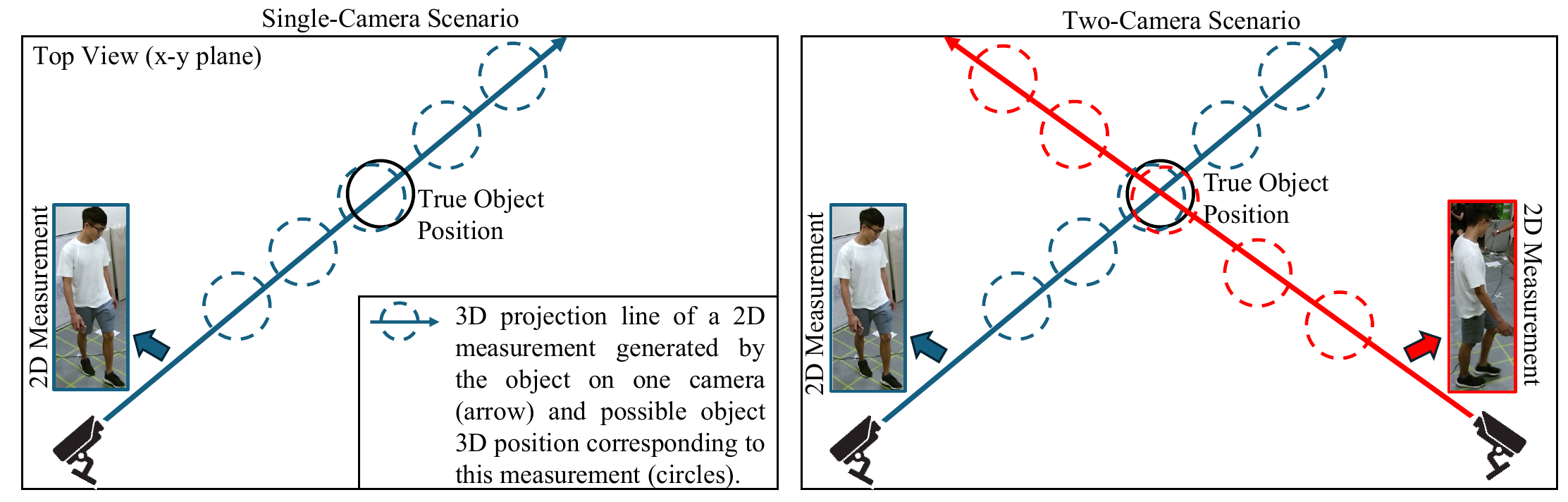}
\par\end{centering}
\caption{An illustration for the requirement of overlapped camera fields of view. Suppose an object is only observed on one camera. In that case, the uncertainty of the 3D object state distribution cannot be resolved along the 3D projection line of the observed 2D measurement, since given a 2D measurement, there are many 3D object states that could generate this measurement with a high likelihood (without any prior information, e.g., 3D training data), hence the high uncertainty in the updated 3D object state distribution. Suppose the object is observed on multiple cameras. In that case, the 3D object state uncertainty is resolved by the complementary information from different views, hence the lower uncertainty in the updated 3D object state distribution.\label{fig:illustration}}
\end{figure*}

\begin{figure}[ht]
\begin{centering}
\includegraphics[width=0.4\textwidth]{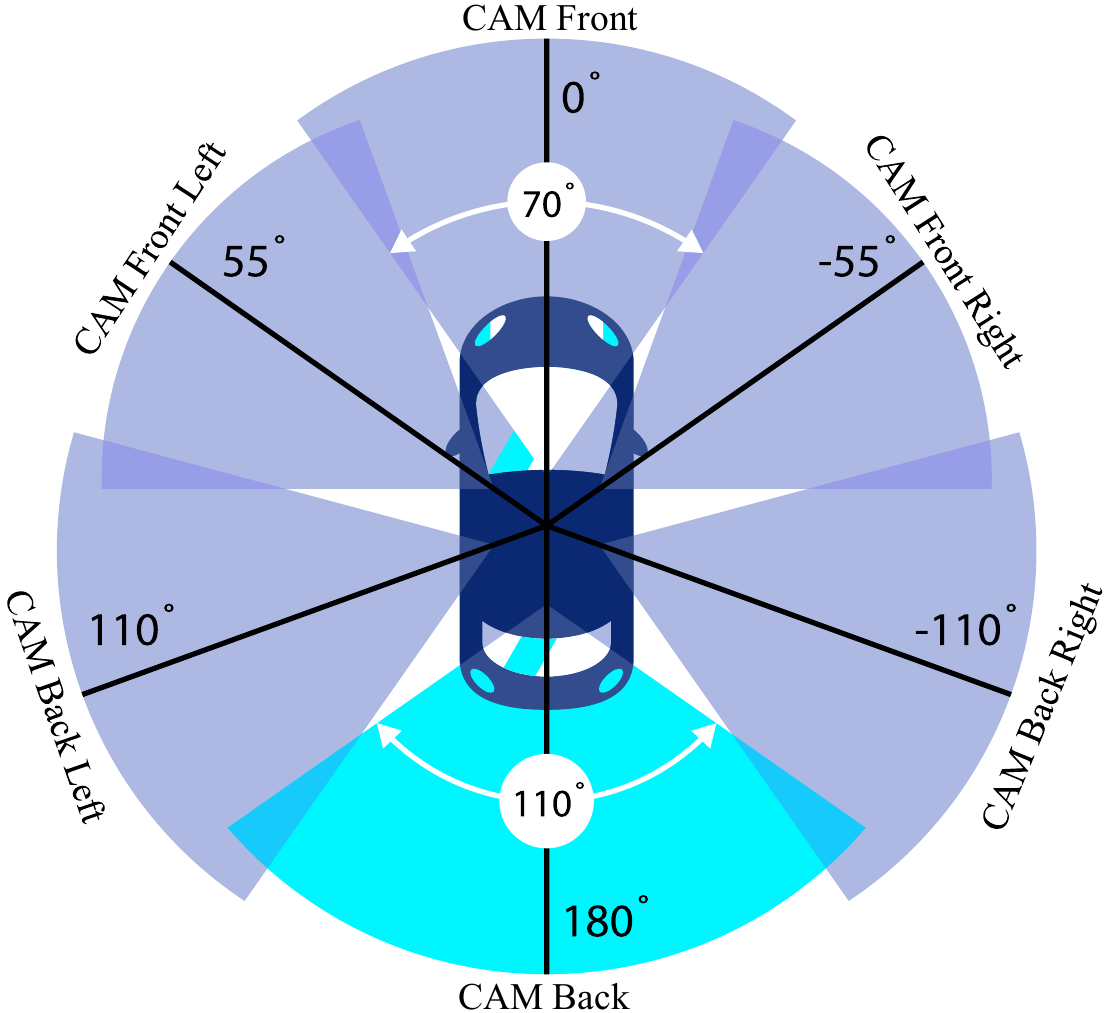}
\par\end{centering}
\caption{Overlapping camera Field of Views in nuScenes dataset (Obtained from \textit{https://scale.com/open-av-datasets/nuscenes}). Cameras are set up to look in six different directions (Front, Front Left, Front Right, Back, Back Left, and Back Right). One camera has very little overlapping FoVs with at most two cameras. The angle of view of CAM Back is 110 degrees, the other five cameras are 70 degrees.
\label{fig:carsensors}}
\end{figure}

Previous methods such as MV-GLMB \cite{ong2020bayesian} only use 2D bounding box detection from monocular cameras to track 3D objects, not using any 3D training data. Their approaches employ multi-sensor MOT to process 2D object detection from multiple cameras with overlapping fields of view to output objects' 3D position and shape. Note that it is impossible to obtain the 3D shape and location of an object with only a single 2D bounding box detection (without 3D training data from the same camera configuration). To accurately update the multi-object density and initialize 3D tracks, we need the fields of view (FoVs) of the cameras to be overlapping to make use of the complementary information from different views, see Fig. \ref{fig:illustration} for our simple illustration. Note that the illustrated phenomenon is well-known in tracking with bearing-only measurements.

In vision-only 3D multi-object tracking, in the CMC dataset \cite{ong2020bayesian}, a person is in the FoVs of at least 3 cameras, and in WILDTRACK dataset \cite{chavdarova2018wildtrack} a person is in the FoVs of at least 4 cameras. Such a setting is standard for 3D surveillance/motion tracking applications applied in MV-GLMB, and their method yields good performance in those standard scenarios. Further, we emphasize that their method does not require 3D training data and the configuration of the cameras can be changed without the need of re-training the detection model. Noting the fact that collecting 3D training data is difficult, training a 3D detection model is expensive, and the model needs to be re-trained if the camera configuration changes.

MV-GLMB based on MS-GLMB framework has low tracking performance on low overlapping field of view such as nuScences dataset \cite{nuscenes2019}. The cameras used in the nuScenes dataset do not have enough overlapping in the FoVs since cameras are installed to look in different directions for coverage, see Fig. \ref{fig:carsensors}. There is virtually no complementary geometric information that our filter can use to perform 3D estimation, analogous to solving a system of 2 simultaneous equations in 3 variables, hence we expect it to have poor performance on the nuScenes dataset (note that MV-GLMB method only processes camera outputs, not the outputs from other sensors). Hence, we present a new implementation method based on MS-GLMB framework that can process camera data and detection from a LiDAR sensor. Specifically, we introduce a novel LiDAR measurement model, which we integrate with an existing multi-camera measurement model to create a new multi-camera and LiDAR multi-object measurement model. These models are incorporated into the MS-GLMB filter to effectively process detections from both LiDAR data and cameras, enhancing the tracking of multiple objects in 3D without the need for overlapping FoVs.


In Section \ref{sec:backgrounds}, we will discuss recent related works employing cameras and LiDAR for 3D multi-object tracking. In Section \ref{sec:camera-lidar-tracking}, we will introduce the measurement and dynamic mathematical models with the Bayes recursion to fuse information from cameras and LiDAR to track 3D objects. In Section \ref{sec:experiment}, we experimentally compare our proposed tracking method with existing approaches on nuScenes dataset. Section \ref{sec:conclusion} summarizes our proposed method and outlines possible directions for future research.

\section{Related Works}\label{sec:backgrounds}
\noindent\textbf{3D LiDAR Based Tracking}.\\
CenterPoint \cite{yin2021center} initially identifies objects' centers using a keypoint detection approach followed by regression to estimate attributes such as 3D dimension size, velocity, and 3D orientation. 3D bounding boxes are fed into a single-camera multi-object tracking algorithm (such as Global Nearest Neighbor (GNN) \cite{blackman1999design}) to output 3D tracklets. The authors in \cite{chiu2001probabilistic} employs Mahalanobis distance and 3D LiDAR detection  MEGVII \cite{zhu2019class} to associate 3D detections to tracks. They further integrate 3D LiDAR point clouds and 2D images to capture the geometric information and appearance of an object  \cite{Chiu2020Probabilistic3M}.

\noindent\textbf{Cameras Based Tracking}.\\
A new high-resolution dataset and large-scale features seven static cameras positioned in a public outdoor space, capturing unscripted dense groups of pedestrians both walking and standing is published in \cite{chavdarova2018wildtrack}, WILDTRACK dataset in which the authors use DeepOcclusion \cite{baque2017deep} with KSP \cite{berclaz2011multiple} and ptrack\cite{maksai2017non} to track objects in 3D. MultiviewX is published in \cite{hou2020multiview}, it has six cameras with high overlapping fields of view. However, this dataset is synthesized. Most methods (for example \cite{teepe2024earlybird,cheng2023rest,nguyen2022lmgp}) use 3D detection such as \cite{qiu20223d} followed by standard multiple-object tracking algorithms to track 3D objects on the ground plane. However, MV-GLMB \cite{ong2020bayesian} with published CMC (Curtin Multi-Camera) dataset is the first method to track 3D objects with width, length, height, and 3D position of objects. Our proposed method is similar to MV-GLMB but we use mean-shift clustering to initiate new birth targets from camera and LiDAR sensors adaptively.

\noindent\textbf{Multi-modality for 3D Multi-object Tracking}.\\
EagerMOT \cite{kim2021eagermot} argued that depth sensors such as LiDAR have a limited sensing range. Hence, they proposed a method that integrates all available information from RGB cameras and depth sensors.  By utilizing images, we can detect distant approaching objects, while depth estimates enable accurate trajectory localization once the objects enter the range of the sensor. Like EagerMOT, our proposed method employs information from depth sensors and cameras to update and initiate new targets.  The authors \cite{wolters2024unleashing} proposed a low-cost, vision-centric 3D perception system that utilizes Radars and Cameras. Among all the modality tracking, the combination of LiDAR and RGB cameras has the highest tracking accuracy according to nuScenes tracking task leaderboard \cite{nuscenes2019}.

\section{Multi-modality based Tracking Using $\delta$-MS-GLMB}\label{sec:camera-lidar-tracking}
\subsection{Object Dynamic Model}\label{subsec:MOT-transition}


Given an object with its state denoted as $\boldsymbol{x}=(x,\ell)$, comprises an attribute $x$ selected from $\mathbb{X}$ (the attribute space), along with a label $\ell$ sourced from $\mathbb{L}$ (a finite label space). Upon the creation of an object at step $k$, it receives a label $\ell=(k,\tau)$ that remains constant over time, where $\tau$ acts as a distinct marker to distinguish between objects that emerge simultaneously in the same step. The attribute $x$ includes the shape parameter $\zeta$ (size of an object), 3D spatial velocity $\dot{\rho}$, and 3D position $\rho$. At any given step $k$, the \textit{multi-object state} is expressed as a \textit{finite collection of distinct object states} within $\mathbb{X}\times\mathbb{L}$, each associated with a unique label \cite{vo2013labeled}.


At step $k$, a collection of newly generated objects is formed. The complete range of potential labels for these objects at step $k$ is denoted as a subset of $\mathbb{L}$, termed $\mathbb{B}$. For a new object assigned the label $\ell$, it is created with a probability of $r_{B}^{(\ell)}$, and given this label, its characteristics are assigned based on distribution $p_{B}^{(\ell)}$. The parameters governing the birth process ${(r_{B}^{(\ell)},p_{B}^{(\ell)})}_{\ell\in\mathbb{B}}$ can either be specified beforehand as a constant value (if we know the prior distribution of newborn objects) or inferred from the data.


At time step $k$, for a particular multi-object state $\boldsymbol{X}$, $(x,\ell)\in\boldsymbol{X}$ has a probability $P_{S,+}(x,\ell)$ of surviving to the subsequent time step or a probability $1-P_{S,+}(x,\ell)$ of dying. If the object persists, it moves to the new state $(x_{+},\ell_{+})$ based on the transition density function $f_{S,+}(x_{+}|x,\ell)\delta_{\ell}[\ell_{+}]$ \cite{vo2013labeled}. Here, the generalized Kronecker delta $\delta_{\ell}[\ell_{+}]$ equals 1 if $\ell$ matches $\ell_{+}$ and is 0 in all other cases, thereby guaranteeing that the label continues to be the same. The state of multiple objects at the subsequent time step, denoted as $\boldsymbol{X}_{+}$, is comprised of both newly generated and existing/surviving objects. This state is governed by the \textit{multi-object Markov transition density} $\boldsymbol{f}_{+}(\boldsymbol{X}_{+}|\boldsymbol{X})$. From now on, the subscript '+' is used to denote the subsequent/next time step.


In this study, we calculate the survival probability by following the model introduced by \cite{kim2019labeled}. The shape parameter, $\zeta$, is defined as a set of three logarithmic values representing the semi-axes of the ellipsoid that encloses the object, following a stochastic process. The kinematic variables, ($\rho$, $\dot{\rho}$), are described using an approach based on approximately constant velocity. For a given $x$, its attribute $x_{+}$ in the subsequent step follows the distribution described in \cite{ong2020bayesian,linh2024inffus}, which can be expressed by the following equation:

\begin{equation}
f_{S,+}(x_{+}|x,\ell)=\mathcal{N}\left(x_{+};Fx+b,Q\right),\label{eq:transition_density}
\end{equation}
where 
\[
F=\left[\begin{array}{cc}
I_{3}(T) & 0_{6\times3}\\
0_{3\times6} & I_{3}
\end{array}\right],I_{3}(T)=I_{3}\otimes\left[\begin{array}{cc}
1 & T\\
0 & 1
\end{array}\right],
\]
\[
b=\left[\begin{array}{c}
0_{6\times1}\\
-\nu^{(\zeta)}/2
\end{array}\right],Q=\left[\begin{array}{cc}
V(\nu^{(\rho)},T) & 0_{6\times3}\\
0_{3\times6} & \textrm{diag}(\nu^{(\zeta)})
\end{array}\right],
\]
\[
V(\nu^{(\rho)},T)=\textrm{diag}(\nu^{(\rho)})\otimes\left[\begin{array}{c}
\frac{T^{2}}{2}\\
T
\end{array}\right]\left[\begin{array}{cc}
\frac{T^{2}}{2} & T\end{array}\right],
\]
the variable $T$ denotes the data collection interval, whereas the three-dimensional vectors $\nu^{(\zeta)}$ and $\nu^{(\rho)}$ indicate the variances of the noise associated with the shape and position parameters, in that order. The logarithmic values of the half-lengths follow a Gaussian distribution, ensuring they remain zero or positive. This indicates that the half-lengths adhere to log-normal distributions, characterized by a unit mean and specific variances $e^{\nu_{i}^{(\zeta)}}-1$, for $i\in\{1,2,3\}$ \cite{ong2020bayesian,linh2024inffus}.


We adopt the multi-camera measurement model presented in \cite{linh2024inffus}, but we define the variables $\nu_{e}^{(c)}$ and $\nu_{p}^{(c)}$ to represent the noise variances for the extent and the center (in logarithmic form) of the bounding box, respectively. We encourage readers to refer to Subsection 3.2 in \cite{linh2024inffus} for more details. In the following section, we will outline our newly proposed LiDAR measurement model.

\subsection{LiDAR Measurement Model}
Given a LiDAR and  multi-object state $\boldsymbol{X}$,  the LiDAR $(l)$ detects $\boldsymbol{x}\in\boldsymbol{X}$ with probability $P_{D}^{\left(l\right)}\left(\boldsymbol{x};\boldsymbol{X}\right)$. Upon detection, it produces a measurement $z^{\left(l\right)}\in\mathbb{Z}^{(l)}$ (where $\mathbb{Z}^{(l)}$ represents the measurement space of LiDAR) with likelihood $g^{\left(l\right)}(z^{\left(l\right)}|\boldsymbol{x})$. Alternatively, the LiDAR may miss-detect $\boldsymbol{x}\in\boldsymbol{X}$ with probability $1-P_{D}^{\left(l\right)}\left(\boldsymbol{x};\boldsymbol{X}\right)$. Similar to camera-based detection, an object detector is required to produce bounding boxes. Measurements obtained from LiDAR only contain 3D points, we refer to measurements here obtained after applying a LiDAR object detection method such as CenterPoint \cite{yin2021center}. 

Similar to the single camera measurement model, the LiDAR observes $\boldsymbol{x}$ as an ellipsoid $z^{(l)}=(z_{p}^{(l)},z_{e}^{(l)})$ where $z_{p}^{(l)}$ is the LiDAR detected object center, $z_{e}^{(l)}$ refers to its size, which is defined by the logarithms of its 3D dimensions: width, length, and height. The measurement set $Z^{(l)} =\{z^{(l)}\}$ includes measurements originating from objects as well as independent false positives, also known as clutter. The LiDAR single-object measurement likelihood  can be written as
\begin{multline}
g^{(l)}(z_{p}^{(l)},z_{e}^{(l)}|x,\ell)=\\
\mathcal{N}\left(\left[\begin{array}{c}
z_{p}^{(l)}\\
z_{e}^{(l)}
\end{array}\right];\textrm{diag}\left(\left[\begin{array}{c}
\nu_{p}^{(l)}\\
\nu_{e}^{(l)}
\end{array}\right]\right)\right),\label{likelihood_lidareq}
\end{multline}
where $\nu_{e}^{(l)}$ and $\nu_{p}^{(l)}$ are the noise variances associated with the extent and the center (in logarithmic scale) of the detected LiDAR ellipsoid, respectively.

At time step $k$, an \textit{association map} for LiDAR is described as a function ${m}^{\left(l\right)}:\mathbb{L}\rightarrow{-1:|Z^{\left(l\right)}|}$, which guarantees that each label corresponds to no more than one measurement. Here, $|Z^{\left(l\right)}|$ denotes the number of elements (cardinality) in the set $Z^{\left(l\right)}$ \cite{vo2013labeled}. For a specific label $\ell$, the notation ${m}^{\left(l\right)}(\ell)=-1$ signifies that the object does not exist, ${m}^{\left(l\right)}(\ell)=0$ represents a missed detection at LiDAR $l$, and ${m}^{\left(l\right)}(\ell)>0$ indicates that the label $\ell$ results in the measurement $z_{{m}^{(l)}\left(\ell\right)}^{(l)}$ captured by LiDAR $l$. Let $\mathcal{L}(\boldsymbol{X})$ refer to the set of labels corresponding to the multi-object state $\boldsymbol{X}$, $\Gamma^{\left(l\right)}$ represent the set of all association maps, and $\mathcal{L}_{{m}^{(l)}}$ $\triangleq\left\{ \ell:{m}^{(l)}\left(\ell\right)\geq0\right\} $ denote the set of active labels ${m}^{(l)}$.  Consequently, the \textit{LiDAR multi-object measurement likelihood} for is defined in \cite{vo2013labeled},

\begin{equation}
\!\!\boldsymbol{g}^{(l)}(Z^{\left(l\right)}|\boldsymbol{X})\propto\!\sum_{{m}^{\left(l\right)}\in\Gamma^{\left(l\right)}}\!\!\!\!\delta_{\mathcal{L}({m}^{\left(l\right)})}[\mathcal{L}\left(\boldsymbol{X}\right)]\!\left[\psi_{Z^{\left(l\right)},\boldsymbol{X}}^{(l,{m}^{\left(l\right)}(\mathcal{L}(\cdot)))\!\!}\left(\cdot\right)\!\right]^{\boldsymbol{X}}\!\!,\label{e:Single_lidar_Lkhd}
\end{equation}
where $\delta_{C}[D]=1$ if $C=D$ and otherwise takes the value of zero, 
\begin{align}
\psi_{\{z_{1:|Z^{\left(l\right)}|}^{(l)}\},\boldsymbol{X}}^{(l,j)}\left(\boldsymbol{x}\right)= & \begin{cases}
\frac{P_{D}^{\left(l\right)}\left(\boldsymbol{x};\boldsymbol{X}\right)g^{\left(l\right)}(z_{j}^{(l)}|\boldsymbol{x})}{\kappa^{\left(l\right)}(z_{j}^{(l)})}, & \!\!\!\!\!j>0\\
1-P_{D}^{\left(l\right)}\left(\boldsymbol{x};\boldsymbol{X}\right), & \!\!\!\!\!j=0
\end{cases},\label{e:lidarPsi}
\end{align}
where LiDAR false positives are typically characterized by an intensity function $\kappa^{\left(l\right)}$.

\subsection{Multi-Camera and LiDAR Multi-Object Measurement Model}

Observing that ${m}^{(1)}(\ell)=...={m}^{(C)}(\ell)={m}^{(l)}(\ell)=-1$ when $\ell$ does not exist, we define a \textit{multi-view association map} ${m}$ as a tuple ${m}\triangleq({m}^{\left(1:C, l\right)})$. $C$ is the number of cameras, $l$ is the index of LiDAR, we only consider one LiDAR in this work. For any $\ell$, ${m}^{(c)}(\ell)=-1$ for one camera $c$, it follows that ${m}^{(c)}(\ell)=-1$ for all available cameras $c$ as well as for LiDAR $l$, $\left({m}^{(l)}(\ell)=-1\right)$. This establishes  ${m}:\mathbb{L}\rightarrow{-1}^{C}\uplus(\mathbb{J}^{(1)}\times\cdots\times\mathbb{J}^{(C)}\times\mathbb{J}^{(l)})$, where $\mathbb{J}^{(c)}\triangleq\{0:\left|Z^{(c)}\right|\}$. Let $\Gamma$ represent the space of multi-view association maps, and let $Z\triangleq\{Z^{\left(1:C\right)}, Z^{\left(l\right)}\}$, we assume these constituent sets are mutually independent conditional on $\boldsymbol{X}$, the \textit{multi-sensor multi-object measurement likelihood} is given by \cite{vo2019multi}:

\begin{equation}
\boldsymbol{g}\left(Z|\boldsymbol{X}\right)\propto\sum_{{m}\in\Gamma}\delta_{\mathcal{L}_{{m}}}[\mathcal{L}\left(\boldsymbol{X}\right)]\left[\psi_{Z,\boldsymbol{X}}^{({m}(\mathcal{L}(\cdot)))}\left(\cdot\right)\right]^{\boldsymbol{X}},\label{e:Multi_Sensor_Lkhd}
\end{equation}
where $\mathcal{L}_{{m}}\triangleq\{\ell:{m}^{(1)}(\ell),...,{m}^{(C)}(\ell),{m}^{(l)}(\ell)\geq0\}$
represents the \textit{live/active label set} of the multi-view association map ${m}$, and
\begin{align}
\psi_{Z,\boldsymbol{X}}^{(j^{(1:C,l)})}\left(\boldsymbol{x}\right) & \triangleq\psi_{Z^{(l)},\boldsymbol{X}}^{(l,j^{(l)})}\left(\boldsymbol{x}\right)\prod\limits _{c=1}^{C}\psi_{Z^{(c)},\boldsymbol{X}}^{(c,j^{(c)})}\left(\boldsymbol{x}\right).\label{eq:Psi-multiSensor}
\end{align}

The Multi-View GLMB recursion, utilizing the two-stage approximation of the Bayes MV-MOT filter, builds upon the work presented in \cite{linh2024inffus}.

\section{Experiments}\label{sec:experiment}

\begin{table*}[h!]
    \centering
    \global\long\def\arraystretch{1.2}%
    \caption{Performance of our proposed algorithm and other state-of-the-art 3D tracking methods on nuScenes validation split.\label{tab:mod-performance}\vspace{1em}}
    \begin{tabular}{|l|c|c|c|c|c|}
    \hline
        Tracker & AMOTA$\uparrow$ & AMOTP $\downarrow$ & RECALL$\uparrow$ & MOTA$\uparrow$ & IDS $\downarrow$ \tabularnewline
        \hline 
        \hline
        Ours & 0.382 & 1.235 & 55.6\% & 0.399 & 2929 \\ \hline
        MV-GLMB-AB \cite{linh2024inffus} & 0.05 & 1.90 & 9\% & 0.08 & 634 \\ \hline
        MUTR3D \cite{zhang2022mutr3d} & 0.294 & 1.498 & 42.7\% & 0.267 & 3822 \\ \hline
        PF-Track-F \cite{pang2023standing} & 0.479 & 1.227 & 59.0\% & 0.435 & 181 \\ \hline
        PermaTrack \cite{tokmakov2021learning} & 0.109 & - & 23.0\% & 0.081 & - \\ \hline
        CenterTrack \cite{zhou2020tracking} & 0.068 & - & 23.0\% & 0.061 & - \\ \hline
    \end{tabular}
\end{table*}

\begin{table*}[!ht]
    \centering
    \global\long\def\arraystretch{1.2}%
    \caption{Performance of our proposed algorithm for various object classes on nuScenes validation split.\label{tab:resultclasses}\vspace{1em}}
    \begin{tabular}{|l|c|c|c|c|c|c|c|c|c|c|c|c|}
    \hline
        Class & AMOTA$\uparrow$ & AMOTP$\downarrow$ & RECALL$\uparrow$ & GT & MOTA$\uparrow$ & MOTP$\downarrow$ & MT$\uparrow$ & ML$\downarrow$ & TP$\downarrow$ & FP $\downarrow$ & FN $\downarrow$ & IDS $\downarrow$ \tabularnewline
        \hline 
        \hline 
        bicycle & 0.249 & 1.489 & 0.372 & 1993 & 0.305 & 0.309 & 15 & 78 & 709 & 102 & 1251 & 33 \\ \hline
        bus & 0.392 & 1.291 & 0.545 & 2112 & 0.451 & 0.487 & 33 & 39 & 1094 & 141 & 960 & 58 \\ \hline
        car & 0.379 & 1.317 & 0.502 & 58317 & 0.433 & 0.38 & 843 & 1612 & 27854 & 2599 & 29030 & 1433 \\ \hline
        motorcy & 0.257 & 1.465 & 0.416 & 1977 & 0.327 & 0.338 & 15 & 51 & 784 & 138 & 1155 & 38 \\ \hline
        pedestr & 0.563 & 0.98 & 0.71 & 25423 & 0.592 & 0.43 & 808 & 358 & 17426 & 2370 & 7373 & 624 \\ \hline
        trailer & 0.23 & 1.456 & 0.442 & 2425 & 0.306 & 0.579 & 32 & 68 & 1038 & 295 & 1353 & 34 \\ \hline
        truck & 0.291 & 1.391 & 0.445 & 9650 & 0.346 & 0.353 & 98 & 263 & 4169 & 833 & 5352 & 129 \\ \hline
    \end{tabular}
\end{table*}

\noindent \textbf{Dataset and Evaluation Metrics}: The nuScenes dataset is specifically developed for applications in autonomous driving using a combination of different sensing modalities including cameras, a LiDAR, and radars.  We evaluate the effectiveness of our algorithm on the NuScenes. Specifically, the NuScenes dataset comprises 1,000 driving sequences, split into 150 sequences each for testing and validation, and 700 sequences for training. Each sequence in the NuScenes dataset spans approximately around 20 seconds, with keyframes extracted at a rate of 2 frames per second. 

We report results from the 150 validation sequences. To compare the performance of our method with existing approaches, we utilize the primary metric/score, Average Multi-Object Tracking Accuracy (AMOTA), and other scores such as Multi-Object Tracking Precision (AMOTP), RECALL, Identity Switch (IDS). These metrics/scores are also the evaluation metrics used in The NuScenes Tracking Challenge \cite{nuscenes2019}.

\noindent \textbf{Parameters}: We consistently set the following parameters throughout all experiments, with the dynamic noise variance (expressed in square meters)  $\nu^{(\zeta)}$=$[0.0036,0.0036,0.0004]^{T}$ and $\nu^{(\rho)}$=$[0.0225,0.0225,0.0225]^{T}$. The noise variance for measurement (bounding box detection, expressed in pixels) in each camera is configured to $\nu_{e}^{(c)}$=$[0.00995,0.0025]^{T}$, $\nu_{p}^{(c)}=[400,400]^{T}$ for pedestrian class and $\nu_{e}^{(c)}$=$[0.0025,0.00995]^{T}$ for car, truck, bus, trailer, motorcycle, bicycle classes. Since the detectors generate a low number of false detections, we use a fixed clutter rate (for applications with higher clutter rates, they can be estimated dynamically from the data).

Previous methods that rely only on cameras, such as the MV-GLMB method \cite{ong2020bayesian}, use fixed birth for target initialization. On the other hand, MV-GLMB-AB \cite{linh2024inffus} initializes new targets from clusters of camera-detected measurements. However, both approaches lack sufficient information about object dimensions such as width, length, and height when initiating tracks from cameras. In the nuScenes dataset, we observed minimal overlap in the fields of view among all the cameras. With limited complementary information from overlapping FoV, these methods are forced to initialize objects with default sizes, which can result in incorrect initialization. This deficiency prevents accurate updates or matching with the correct 2D camera-detected bounding boxes, leading to false targets. Hence, birth targets generated from camera data are susceptible to false positive tracks. Since LiDAR detections include information about object size, we rely exclusively on LiDAR 3D detection data to establish new tracks. We observe that if an object is beyond the LiDAR detection range, it is likely absent from the ground truth. Furthermore, When a detection/measurement is less likely linked with existing tracks, it probably originates from a new birth. Hence, we generate new birth tracks based on the adaptive birth model described in  \cite{reuter2014labeled}, wherein measurements from the current time step are employed to initialize tracks in the subsequent time step.

There are seven classes in nuScenes dataset including pedestrian, car, truck, bus, trailer, motorcycle, and bicycle. For different class, we set different noise variances for the detected LiDAR ellipsoid, denoted as $\nu^{(l)} = \{\nu_{p}^{(l)}, \nu_{e}^{(l)}\}$. These variances are determined based on each class's real average object size, where larger object dimensions (width, length, height) have higher variances. The specific settings are as follows: \{\textit{pedestrian}: [0.1 , 0.1 , 0.1 , 0.005, 0.005, 0.005], \textit{car}: [2.0 , 2.0 , 2.0 , 0.405, 0.405, 0.405], \textit{truck}: [2.0 , 2.0 , 2.0 , 0.405, 0.405, 0.405], \textit{bus}: [2.0 , 2.0 , 2.0 , 0.405, 0.405, 0.405], \textit{trailer}: [2.0 , 2.0 , 2.0 , 0.405, 0.405, 0.405], \textit{motorcycle}: [0.5 , 0.5 , 0.5 , 0.005, 0.405, 0.005], \textit{bicycle}: [0.5 , 0.5 , 0.5 , 0.005, 0.405, 0.005]\}. We process/track measurements for each class independently to prevent incorrect associations. For instance, a motorcycle measurement might be erroneously assigned to bicycle tracks. To improve our 3D LiDAR detection system, we only select detections with a confidence score greater than 0.47. However, we could consider employing the adaptive confidence score method for LiDAR detection as described by \cite{van2023adaptive}, though this approach requires further investigation.

We use CenterPoint \cite{yin2021center} LiDAR detection model to obtain 3D measurements from the 3D point cloud. This detector outputs 3D bounding box measurements with center, width, length, and height. In our filter, we use a Gaussian (single-object) likelihood function for the 3D measurements observed by the LiDAR sensor, and assume the detection process of the cameras and LiDAR are independent. We report the performance of our algorithm and other existing approach methods in Table \ref{tab:mod-performance}. 

We observe that the performance of our method is on par with the MUTR3D \cite{zhang2022mutr3d} algorithm but falls short compared to the PF-Track-F \cite{pang2023standing} algorithm. However, our method surpasses the PermaTrack \cite{tokmakov2021learning} and CenterTrack \cite{zhou2020tracking} algorithms, both of which also rely on 3D training data. Compared to the two filters in the MS-GLMB framework, our method outperforms the MV-GLMB-AB \cite{linh2024inffus}. Notably, the MV-GLMB-AB has a lower number of identity switches than our method. This is because MV-GLMB-AB, relying solely on camera data, lacks sufficient information to update the object state. As a result, these birth targets quickly disappear after moving out of the camera's FoV leading to tracking a few numbers of objects. Due to its fixed birth procedure, we could not replicate the results for MV-GLMB \cite{ong2020bayesian}.

Table \ref{tab:resultclasses} shows the tracking results across various classes. Pedestrian tracking is the most effective, achieving the highest AMOTA of 0.563, but it also has significant false positives (2370) and false negatives (7373), indicating challenges in distinguishing pedestrians. In contrast, Bicycles and Trailers show the lowest AMOTA values (0.249 and 0.230, respectively), suggesting poorer tracking performance. Additionally, the Car class experiences the highest identity switches (1433), highest false negatives (29030), and highest (2599) signaling issues with maintaining a consistent identity. We believe that our initial parameters, such as these parameters $\nu^{(\rho)}, \nu^{(\zeta)}$ (position and shape noise variances), $\nu_{e}^{(c)}, \nu_{p}^{(c)}$ (cameras' detection bounding boxes noise variances), and $\nu_{e}^{(l)}, \nu_{p}^{(l)}$ (the corresponding parameters for LiDAR detections) for the Car class, may not have been optimally set. Therefore, tuning these parameters represents a potential avenue for future work aimed at improving accuracy.

\section{Conclusion}\label{sec:conclusion}
The proposed enhancement of the MS-GLMB-based filters through the incorporation of an additional sensor, such as LiDAR, has demonstrated significant improvements in multi-object tracking performance. This advancement effectively addresses the limitations of existing works, which rely on overlapping camera fields of view. To further advance the robustness and versatility of multi-object tracking systems, future research should explore the integration of additional sensing modalities, such as radar. Additionally, efforts should be directed toward minimizing the dependency on overlapping FoVs, and parameter initialization enabling more flexible and comprehensive tracking solutions across diverse and complex environments.


\bibliographystyle{IEEEtran}
\bibliography{reflib_abb}

\end{document}